\documentclass[]{OAGM}
\OAGMarXiv{1304.1876}

\usepackage{graphicx}

\usepackage{amsmath}
\usepackage{amssymb}
\usepackage{latexsym}

\DeclareMathOperator*{\argmin}{arg\,min}

\long\def\symbolfootnote[#1]#2{\begingroup\def\thefootnote{\fnsymbol{footnote}}\footnote[#1]{#2}\endgroup} 

\usepackage[tight]{subfigure}

\usepackage{framed}

\graphicspath{{../code/matlab/out/}{img/}{./}}

\title{A Convex Approach for Image Hallucination}
\author{Peter Innerhofer, Thomas Pock\\
Institute for Computer Graphics and Vision, University of Technology Graz
}

\begin{document}
\maketitle

\begin{abstract}
  In this paper we propose a global convex approach for image hallucination. Altering the idea of classical multi image
super resolution (SU) systems to single image SU, we incorporate aligned images to hallucinate the output. Our work is based on
the paper of Tappen et al.\cite{tappen_bayesian_2012} where they use a non-convex model for image hallucination. In comparison we
formulate a convex primal optimization problem and derive a fast converging primal-dual algorithm with a global optimal 
solution. We use a database with face images to incorporate high-frequency details to the high-resolution output. We show that we
can achieve state-of-the-art results by using a convex approach.
\end{abstract}

\section{Introduction}
\label{sec:introduction}

Single image Super Resolution (SU) systems yield to estimate a high-resolution (HR) image from low-resolution (LR) input. This is clearly
an ill-posed problem due to the fact that important high frequency information is lost in a down-sampling process. 

A common constraint to nearly all SU systems is the reconstruction constraint, which says that the HR result down-sampled
 should be the same as the LR input. However, this constraint is weak and the space of possible solutions is
large. A generic smoothness prior, like the Total Variation (TV), can improve this constraint but no lost information is infered.

More advanced systems model edge statistics which can produce HR images with sharp edges while leaving other regions
smooth\cite{sun_sketch_2003}. This approach has its advantages in creating sharp edges with minimal jaggy or squary artifacts.
But their performance will decrease as the resolution of the input decreases because the perceptual important edges will
vanish. Additionally such systems cannot introduce novel high frequency details which were lost in the down-sampling process.

Backer and Kanade have shown in \cite{baker_limits_2002} that systems which only rely on the reconstruction constraint (and possible altered with a smoothness prior) cannot create high frequency image content. They propose the technique of image hallucination where
HR image details and there LR correspondences are learned on a patch basis to synthesise HR images. 
Such systems like~\cite{liu_2001} can introduce new details which are not present in the low resolution image. However, the
patch-selection process remains a key problem in such systems and the mathematical models make it difficult to control artefacts
in the output. A state-of-the-art enhancement of such a system is the work of Sun et al. \cite{sun_context-constrained_2010} which
incorporates their so called textual context bridging the gap between image hallucination and texture synthesis.

While these systems perform well on general images, domain based SU system where the content of the image is known have shown 
improvements of the results again. An example of such an approach is the work of Liu et al. face
hallucination\cite{liu_face_2007}. Their system inferes regularities of face-appearances to hallucinate details that a general image
model can't create. However, the system of \cite{liu_face_2007} is limited to frontal face images and can't handle large pose and
viewpoint changes.

Our work is based on the paper of Tappen el al. \cite{tappen_bayesian_2012}. This approach uses aligned face images prior to the
hallucination process and therefore incorporates the ideas of the classic multi-image SU-systems. Tappen et al.
uses Patch Match\cite{barnes_patchmatch_2009} to quickly search for similar face images in a large database. The best matches are called
candidates. These candidates are densely aligned using the SIFTflow algorithm\cite{liu_sift_2011}. Their system incorporates
a edge focusing image prior, a global likelihood function (the reconstruction constraint) and an example-based non-convex hallucination model within a Baysian framework. Tappen et al. pointed out that if their system can't find good candidates, the performance decreases fast and the results get blurry. These limitations could be compensated by falling back to an edge-based system. Our system tries to improve this behaviour by using a hallucination model more robust to outliers.

\begin{figure}
 \centering
 \includegraphics[width=1\textwidth]{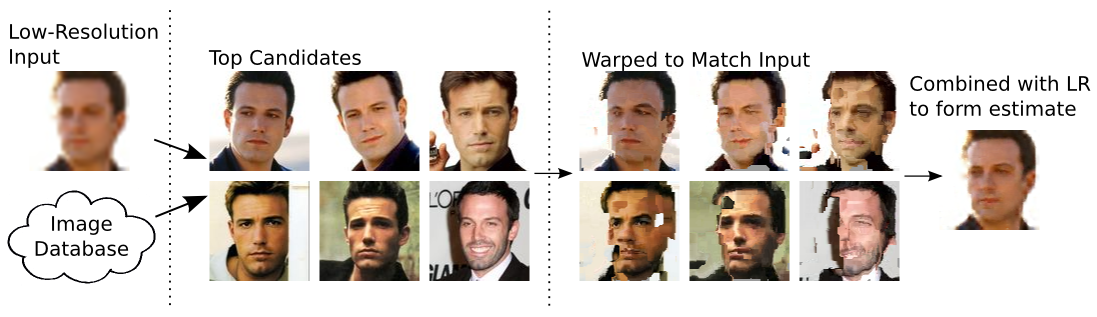}
 \label{fig:system}
 \caption{System-Overview: We use SiftFlow\cite{liu_sift_2011} to find similar appearing images based on the alignment energy. These Canididates are warped to match the input. The LR input and the aligned candidates are incorporated to form the estimate.}
\end{figure}

We proposes a similar work flow, but in comparison all our models are convex and we omit PatchMatch. We search the database with SiftFlow utilizing the SiftFlow energy and warp the search results with the same algorithm. We omit PatchMatch because we think it is more important to have good aligned candidates rather than similar appearing images. Our convex optimization problem joins an total variation based image prior, the reconstruction constraint and a hallucination model robust to outliers. Starting with the primal minimization problem we derive a generic saddle-point problem and solve it with a fast converging primal-dual algorithm proposed by Chambolle et al. \cite{chambolle2011}. Figure \ref{fig:system} shows our system overview.

This paper is organized as followed. After presenting a convex approach for image hallucination in
section~\ref{sec:hallucination} we derive a generic saddle-point problem and solve it with a so called primal-dual algorithm in
section~\ref{sec:algorithm}. In section~\ref{sec:experiments} we describe the experiments made and we conclude in
section~\ref{sec:conclusion}.

\section{A Convex Approach for Image Hallucination}
\label{sec:hallucination}

As pointed out in the introduction, our work alters the model of Tappen et al. \cite{tappen_bayesian_2012} to a convex approach. Solving a convex minimization problem has nice advantages. Convexity guarantees an existing, unique solution\cite{boyd_convex_2004} and fast convergence can be achieved. However, choosing the image models and energy minimization terms are crucial for a perceptual good solution. The minimization problem of \eqref{equ:energy} combines 3 different image models and constraints

\begin{equation} \label{equ:energy}
 u^{\ast} = \argmin_{u}  \underbrace{\|\nabla u\|_{2,1}}_{1} + \underbrace{\lambda\|DBu-f\|^2_2}_{2} +
\underbrace{\gamma\sum_i\| H (u - g_{C_i})\|_1}_{3},
\end{equation}
where \( u \) is our HR estimate, \( f \) the LR input and \( g_C \) are the aligned candidate images.

The first term is a general smoothness prior equipped with the TV-norm. This model preserves sharp edges while staying smooth in
other regions. The TV is defined as $ TV(u) = \int_\Omega |\nabla u|dx $, where  $ \nabla = \left[ \frac{\partial}{\partial
x},\frac{\partial}{\partial y} \right] $ is the gradient operator and the $ \|\cdot\| $ is the L1-norm. 

The second term of \eqref{equ:energy} models the reconstruction constraint. The constraint ensures that the down-sampled
HR image yield to the LR input. In other words, the HR estimate down-sampled should be the same as the input. The matrices $ DB $ composes of a Gaussian blurring or anti-alias filter
$ B $ and a down-sampling matrix $ D $. The reconstruction constraint implies a linear model where the observed image \(f \) is a linear
combination of the undistorted image \(u \) added by noise $ f = Au + n $. If we model the noise as Gaussian, we end up by equipping
this term with a quadratic norm minimizing the noise. The factor $ \lambda $ controls how strong the constraint is imposed.

The third term of \eqref{equ:energy} represents a non-parametric image model here referred as the hallucination term. Having
found similar candidate images from a database and aligned them to our input, high-frequency details can be introduced from these
candidates. The term minimizes the difference between the HR result $ u $ and the candidate images $ g_{C} $
after applying a high-pass filter $ H $. We apply the high-pass filter $ H $ to infer only
high-frequency information from the candidates. Thats because the low-frequency details are still present in the LR input and can be omitted. Equipping this function with the L1-norm makes it
robust to outliers which is the case if no good candidates where found or if the alignment fails.

Note that there exists a strong relation between the blurring matrix \( B \), the high-pass filter \(H\) and the scaling factor. In
fact the high-pass filter kernel equals a all-pass kernel \( \delta \) subtracted by the blurring kernel. So we incorporate just frequencies we lost in the down-sampling process. Moreover the blurring kernel depends on the scaling factor\cite{unger_su_2010}. We choose the standard derivation for the blurring kernel as \( \sigma = \frac{1}{4}\sqrt{\xi^2-1} \), where \( \xi \) is the scaling factor.


\section{Deriving the Primal-Dual Algorithm}
\label{sec:algorithm}

In this section we will derive the first order primal-dual form of \eqref{equ:energy}. We will solve this generic
saddle-point problem with a variational approach, the primal-dual algorithm of Chambolle et al. \cite{chambolle2011}. 

%
The goal is to transform the primal minimization problem \eqref{equ:energy} into a convex-concave saddle-point problem of the type:

\begin{equation}
 \min_x\max_y \left\langle{Kx,y}\right\rangle\ + G(x) -F^\ast(y),
\label{equ:saddlepoint}
\end{equation}
with a continuous linear operator \(K\), and \(G(x)\) and \(F(x)\) being convex functions. 

In a first step one has to apply the Legendre-Fenchel transformation also refered as the conjugate of a function\cite{boyd_convex_2004}. We derive the conjugate of the total variation (TV) \( \|\nabla u\|_{2,1} \) and of the hallucination term \( \gamma\sum_i\|H(u - g_i)\|_1 \) introducing the dual variables \(p\) and \(r_i\) respectively. Additionally the primal variables \(w_i\) is introduced as a lagrange-multipliere of the hallucination term leading to:

\begin{align}
  \min_{u,w \in X } \max_{p,r \in Y} \left\langle{Kx,y}\right\rangle\ + \underbrace{\lambda\|DBu-f\|_2^2 +
\gamma\sum_i\|w_i\|_1}_{G(x)} + \underbrace{\sum_i\left\langle{-H g_i,r_i}\right\rangle\ - \delta_{\|p\|_\infty \leq
1}(p)}_{-F^\ast(y)},
\label{equ:saddlepoint_u}
\end{align}
with the structure of \( K\), \( x\) and \( y\) as:
\begin{equation}
K = 
 \begin{pmatrix}
	 \nabla &     	&   	&		& \\
	 H		&   -I  	&   	&		& \\
	 H  		&       	& -I	&      	& \\
     \vdots	&       	&   	& \ddots & 
 \end{pmatrix}
, \quad x =
 \begin{pmatrix}
   u \\
   w_1\\
   \vdots \\
   w_n 
 \end{pmatrix} 
, \quad y =
 \begin{pmatrix}
   p \\
   r_1\\
   \vdots \\
   r_n 
 \end{pmatrix}.
\label{equ:k}
\end{equation}
The term \( \delta_{\|p\|_\infty \leq 1}(p) \) denotes the indicator function and \(\|p\|_\infty \) the maximum norm.

Note that we don't apply the Legendre-Fenchel transformation to the reconstruction constraint \(\lambda\|DBu-f\|_2^2 \).
Instead we solve this sub-problem using the conjugate-gradient method (CG) \cite{barrett_templates_1994} in an subroutine. Because
the reconstruction constraint imply a linear model and thus consists of a linear system of equations, it is reasonable to use a
fast-converging solver specialized on such systems. We refer to \ref{sec:prox} for further details. 

\subsection{Algorithm}
\label{sec:pd-algorithm}

We use the first order primal-dual algorithm proposed in \cite{chambolle2011}, there referred as ``Algorithm 1''. The idea is to
perform a gradient ascent/decent step on the unconstraint objective function and sequentially reproject the variables according to
the constraints. The gradient step-size \(\sigma\) and \( \tau \) are crucial for convergence and have to satisfy \( \tau\lambda
L^2 < 1 \) with \( L = \|K\| \) the operator norm of \(K \). Within an iteration we perform a gradient decent in the primal
variable \( x \) and a gradient ascent in the dual variable \( y \) followed by the reprojection utilizing the prox-operators.
Additionally we perform a linear extrapolation of the dual variable based on the current and the previous iterates with \( \theta
= 1 \). This can be seen as an approximate extragradient step and offers fast convergence. 

\begin{framed}
\begin{itemize}
 \item Initialization: Let \( \tau\sigma L^2 < 1 \), with \( L = \|K \|, \quad \theta \in [0,1], \quad (x^0,y^0) \in X\times Y \)
and set \(\bar{y} = y^0 \)
 \item Iterations \( (n \geq 0)\): Update \( x^n,y^n,\bar{y}^n\) as follows: \\
\begin{equation}
\left\lbrace
 \begin{array}{l}
	x^{n+1} = (I + \tau \partial G)^{-1}(x^n - \tau K^\ast \bar{y}^{n}) \\
	y^{n+1} = (I + \sigma \partial F^\ast)^{-1}(y^n + \sigma K x^{n+1}) \\
	\bar{y}^{n+1} = (y^{n+1} + \theta (y^{n+1} - y^n)) \\
 \end{array}\right.
 \label{equ:algorithm}
\end{equation}
\end{itemize} 
\end{framed}

\subsection{The prox-operators}
\label{sec:prox}

Proximity Operators are a powerful generalization of projection operators. Their importance is attached by splitting the subject
to be minimized into simpler functions that can be handled individually \cite{combettes2011}. The proximity operator then assures
to ``resolve'' the sub-gradient \( \partial G \) of any function  \( G \) even if \( G \) is non-smooth. We assume that F and G
are simple so that one can compute their proximity operator in a closed-form. The operator is defined as: 
\begin{equation}
 x = (I + \tau \partial G)^{-1}(x) = \argmin_x \left\lbrace \frac{\|x-y\|^2}{2\tau} + G(x) \right\rbrace.
\label{equ:resolvent}
\end{equation}
In order to apply the algorithm we have to compute the prox-operator for \( (I + \sigma \partial F^\ast)^{-1} \) and \( (I + \tau
\partial G)^{-1} \). In \eqref{equ:saddlepoint_u} we see that 

\begin{equation}
 F^\ast(y) = \delta_P - \sum_i\left\langle{-H g_i,r_i}\right\rangle\,
\end{equation}
and 
\begin{equation}
G(x) = \lambda\|DBu-f\|_2^2 + \gamma\sum_i\|w_i\|_1 .
\end{equation}
The first term in \( F^\ast(y) \) is the indicator function of a convex set and the prox- or resolvent operator reduces to a pointwise Euclidean projection onto \( L^2 \) balls. The function \(\left\langle{-H g_i,r_i}\right\rangle\ \) poses an inner product and the prox  operator of reduces to an affine function. 

\begin{equation}
 y = (I + \sigma \partial F^\ast)^{-1}(\tilde{y}) \iff p = \frac{\tilde{p}}{\max(1,\|p\|_{2,1})}, \quad r_i = \tilde{r}_i + \sigma
H g_i 
\end{equation}
For \(\|w_i\|_1 \) the resolvent operator poses a soft-threshold shrinkage function. The prox-operator of the reconstruction
constraint poses again a linear problem: \(A_{new}u = b_{new} \), with \( A_{new} = (I + \lambda\tau A^TA) \) and \(b_{new} =  \lambda\tau A^Tf+\tilde{u} \)
\begin{equation}
\begin{array}{l l}
 x = (I + \tau \partial G)^{-1}(\tilde{x}) & \iff w_i = \left\{
  \begin{array}{l l}
	\tilde{w_i} - \tau\sigma 	& \quad \text{if $\tilde{w}_i > 0$} \\
	\tilde{w_i} + \tau\sigma		& \quad \text{if $\tilde{w}_i < 0$} \\
    0			 				& \quad \text{else}
  \end{array} \right. \\
 & \\
 & \iff (I + \lambda\tau A^TA)(u) = \lambda\tau A^Tf+\tilde{u}
\end{array}
\end{equation}
Note that the CG-method expects a symmetric positive definite matrix \( A_{new} \) which is clearly the case. We apply the CG
with a so called ``hot-start'' where the previous iterate of \( u \) is used for initialization. The hot-start initialization
achieves faster convergence of the CG-method.


\section{Experiments}
\label{sec:experiments}

\begin{figure}[ht]\centering
\subfigure[input]{
	\includegraphics[scale=0.45]{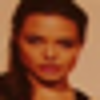}
  }
  \subfigure[our's]{
	\includegraphics[scale=0.45]{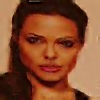}
  } 
  \subfigure[original]{
	\includegraphics[scale=0.45]{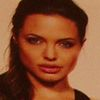}
  } 
  \subfigure[input]{
	\includegraphics[scale=0.45]{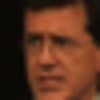}
  }
  \subfigure[our's]{
	\includegraphics[scale=0.45]{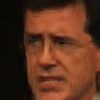}
  } 
  \subfigure[original]{
	\includegraphics[scale=0.45]{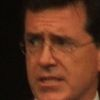}
  }\\
  \subfigure[input]{
	\includegraphics[scale=0.45]{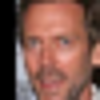}
  }
  \subfigure[our's]{
	\includegraphics[scale=0.45]{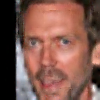}
  } 
  \subfigure[original]{
	\includegraphics[scale=0.45]{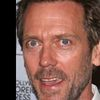}
  }
    \subfigure[input]{
	\includegraphics[scale=0.45]{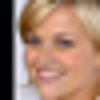}
  }
  \subfigure[our's]{
	\includegraphics[scale=0.45]{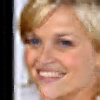}
  } 
  \subfigure[original]{
	\includegraphics[scale=0.45]{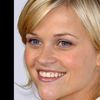}
  }\\
 \label{fig:hallucination}
 \caption{Result of our convex approach with a scaling factor of four. In each image group we present the LR input, our estimate and the original HR version.}
\end{figure}

In our experiments we use the PubFig83 database presented in \cite{pinto_2011}. The database consists of over 14,000 images of
public figures, cropped to include just the faces and resized to the identical resolution of \(100\times100\) pixels. 

All results are produced in the same manner. First we down-sample the input by a factor of 4 using bicubic interpolation, followed
by a bicubic up-sampling by the same factor. We use the resampled image as an input to the SiftFlow \cite{liu_sift_2011} and
search for candidates with the least SiftFlow energy. Figure \ref{fig:system} demonstrates this process. We just search in the set of
pictures from the same individual as the input. We imply that the person of the input has already been identified by a
face-recognition system and pictures of this person are available. Having found the best 6 candidate images, we
aligned them to our input using again Siftflow. With this aligned candidates we run the primal-dual algorithm. Note that the input
image $ f $ of the algorithm is still the bicubic down-sampled \(
25\times25 \) image, while the candidates \( g_i \) and the result \( u^\ast \) are \( 100\times100 \). On the output we calculate
the Signal-to-Noise ratio (PSNR) and the SSIM index. Figure \ref{fig:hallucination}
shows some results of our algorithm.

In our experiments we discovered that a strong reconstruction constraint is needed and therefore the $ \lambda $-value was set to 
$\lambda = 5\cdot10^4 $, which has proven a high PSNR. The hallucination parameter \(\gamma \) was set to \(\gamma = 20 \) so that smoothing by the TV-regulatization is still applied. To treat the color images in optimization, we did a so called channel-by-channel optimization. A more comprehensive color treatment was proposed in \cite{goldluecke_2010} called vectorial total variation. This advanced TV-regulatization should be included in future work.

\begin{figure}[ht]\centering
\subfigure[]{
	\includegraphics[scale=0.39]{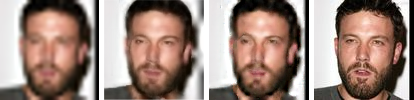}
  }
  \subfigure[]{
	\includegraphics[scale=0.39]{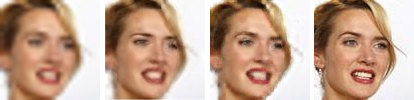}
  }\\
  \subfigure[]{
	\includegraphics[scale=0.39]{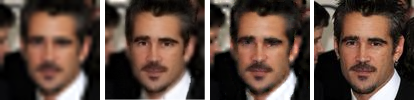}
  }
  \subfigure[]{
	\includegraphics[scale=0.39]{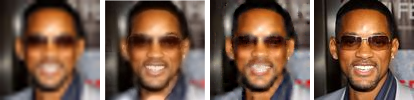}
  }
 \label{fig:comparison}
 \caption{Comparison between the estimates of Tappen et al. and our approach. The first image of each set shows the LR input. In the second image we see the approach of Tappen et al. and the third shows our estimate compared to the actual HR image as fouth.}
\end{figure}

We ran our algorithm on all 14,000 images and got a mean PSNR of 24,13dB. This result outperforms the work of Tappen et al. which got a mean PSNR of 24.05dB. Table \ref{tab:psnr} shows a comparison with different algorithms and the achieved PSNR and SSIM index. The table was partly taken from Tappen et al. and we refer to \cite{tappen_bayesian_2012} for further information. In figure \ref{fig:comparison} we show a comparison between the results of Tappen et al. and our approach. The percetual differences on these results are quite low which is not extraordinary because all these examples achieve a high PSNR and SSIM compared to the average.

\begin{table}[ht!]
\centering
\begin{tabular}{| l | l | l |}
\hline
Algorithm & PSNR (dB) & SSIM Index \\
\hline
VISTA & 23.47 & 0.669 \\
Sun et al. \cite{sun_context-constrained_2010} & 23.82 & 0.741 \\
Tappen et al. \cite{tappen_bayesian_2012} & 24.05 & 0.748 \\
Our Approach & 24,13 & 0.750 \\
\hline
\end{tabular}
\label{tab:psnr}
\caption{Comparison of different algorithms and their achieved PSNR on the PubFig83 database }
\end{table}

\section{Conclusion}
\label{sec:conclusion}

We presented a convex and global approach for image hallucination. This implies a fast converging algorithm with a unique
solution. By incorporating high-frequency information from similar images we get perceptually good solutions. Especially if the
alignment of the candidates image works well, the results can be nearly perfect. A crucial part in our system poses the SiftFlow algorithm, first because we use it as a searching tool, and second and more important we use it for the alignment of the images. If SiftFlow is able to align the images, the results are superior to those where the alignment fails. Tracking failed alignments and replacing such candidates should achieve improvements in future work. 

We think that it is not so important to take images from the same person rather than having good alignments. For future work we propose to build a bag of visual words taken from face images and to apply the same algorithm so that no face-recognition system is needed. 
Due to the fine modeling of the down-sampling, blurring and highpass filter and the robust hallucination model our convex approach achieves good performance and state-of-the-art results.

\section*{Acknowledgments}

This work was supported by the Austrian Science Fund (project no. P22492)

\bibliography{report}
\end{document}